\documentclass{article}

\usepackage[preprint]{neurips_2024}

\usepackage[utf8]{inputenc} 
\usepackage[T1]{fontenc}    

\usepackage[pagebackref,breaklinks,colorlinks]{hyperref} 

\usepackage{url}            
\usepackage{booktabs}       
\usepackage{amsfonts}       
\usepackage{nicefrac}       
\usepackage{microtype}      
\usepackage{xcolor}         

\usepackage{amsmath}
\usepackage{subcaption}
\usepackage{multirow}   
\usepackage{makecell}   
\usepackage{wrapfig}    
\usepackage{scalerel,stackengine}
\stackMath
\newcommand\reallywideacute[1]{%
  \ThisStyle{\savestack\tmpA{$\SavedStyle#1$}%
  \savestack{\tmpbox}{%
    \def\scriptstyleScaleFactor{0.8}
    \def\scriptscriptstyleScaleFactor{0.68}
    $\SavedStyle\stretchto{%
    \scalerel*[\wd\tmpAcontent]%
      {\kern-2.05\LMpt\mathchar"7013\kern-1.1\LMpt}%
    {\rule{0ex}{\textheight}}%
  }{2.0\LMex}$}%
  \stackengine{-7\LMpt}{\SavedStyle#1}{\tmpbox}{O}{c}{F}{T}{S}}%
}

\usepackage{listings}

\usepackage{verbatim}

\title{Adaptive Explicit Knowledge Transfer for Knowledge Distillation}

\author{%
  Hyungkeun Park \\
  Yonsei University \\
  \texttt{hyungkeun.park@yonsei.ac.kr}
  \And
  Jong-Seok Lee \\
  Yonsei University \\
  \texttt{jong-seok.lee@yonsei.ac.kr}
}

\begin{document}

\maketitle

\begin{abstract}
Logit-based knowledge distillation (KD) for classification is cost-efficient compared to feature-based KD but often subject to inferior performance. 
Recently, it was shown that the performance of logit-based KD can be improved by effectively delivering the probability distribution for the non-target classes from the teacher model, which is known as `implicit (dark) knowledge', to the student model. 
Through gradient analysis, we first show that this actually has an effect of adaptively controlling the learning of implicit knowledge. 
Then, we propose a new loss that enables the student to learn explicit knowledge (i.e., the teacher's confidence about the target class) along with implicit knowledge in an adaptive manner. 
Furthermore, we propose to separate the classification and distillation tasks for effective distillation and inter-class relationship modeling. 
Experimental results demonstrate that the proposed method, called adaptive explicit knowledge transfer (AEKT) method, achieves improved performance compared to the state-of-the-art KD methods on the CIFAR-100 and ImageNet datasets.

\end{abstract}

\section{Introduction}
\label{sec:intro}

With the advances of deep learning techniques, the classification performance of deep neural networks has been significantly improved, but often at the cost of increased computational and memory requirements. 
This trend poses challenges for deployment of deep neural networks in resource-constrained environments such as edge devices. 
As a way to resolve this issue, knowledge distillation (KD) has emerged, which aims to transfer knowledge from a trained large model (denoted as teacher) to a smaller model (denoted as student) \cite{gou21knowledge}.
It has been shown that the student model produced by KD can outperform the same-sized model directly trained from scratch \cite{hinton14distilling,dao21knowledge}.

KD methods can be divided into two groups according to the learning approach: feature-based and logit-based. 
In feature-based KD, the student model is trained to mimic the features learned at intermediate layers of the teacher model \cite{romero15fitnet,zagoruyko17paying,heo19comprehensive,park19relational,zhang20task,tian20contrastive,yang21knowledge,huang21revisiting,chen21distilling}. It has an advantage in terms of the performance, but is subject to complexities in layer selection and knowledge transfer mechanisms. Logit-based KD aims to align the probability distributions between the teacher and student models \cite{huang22knowledge,zhao22decoupled,yang23from,hao2023oneforall}. This provides a more straightforward and cost-efficient approach in comparison to feature-based KD.

In the classical logit-based KD, the Kullback-Leibler (KL) divergence between the output probability distributions of the teacher and the student is used as the loss to be minimized \cite{hinton14distilling}. 
In \cite{zhao22decoupled}, it is found that this loss can be reformulated as a weighted sum of the loss for the target class and the loss for the non-target classes, and the weight given to the latter is a factor disturbing effective distillation. Based on this analysis, \cite{zhao22decoupled} proposes a method called decoupled knowledge distillation (DKD) by changing the weight for the second loss term, which is shown to improve the classification performance.

In this paper, we first analyze the classical KD and DKD in a new perspective by examining the gradients backpropagated to the student model's logits during training. 
We find that DKD involves an additional term adjusted by the ratio of the probabilities for the non-target classes from the teacher and the student in the gradients with respect to the non-target logits. 
That is, the performance enhancement by DKD can be attributed to adaptively adjusted learning of the rich information in the non-target logits, which is called `implicit knowledge' (also known as dark knowledge)\footnote{Although the term `dark knowledge' is more popular in literature, we use the term `implicit knowledge' in order to contrast it with `explicit knowledge'.}.

Then, we propose a novel KD method called adaptive explicit knowledge transfer (AEKT), which additionally enables effective learning of `explicit knowledge' (i.e., information contained in the target logit) in an adaptive manner. 
We design a new loss function that can adaptively control the gradient for the target class according to the confidence ratio of the teacher and student models.
Furthermore, we identify the ineffectiveness of enforcing the logits of the student model to minimize both the cross-entropy loss for classification and the distillation loss at the same time.
Therefore, we propose to serialize the classification task and the distillation task using a fully-connected (FC) layer, which facilitates an effective distillation process and helps exploit inter-class relationship.
Experimental results demonstrate that the proposed method outperforms existing state-of-the-art feature-based and logit-based KD methods on the CIFAR-100 and ImageNet datasets.

Our contributions can be summarized as follows.
\begin{itemize}
\item We perform mathematical analysis of KD and DKD in a new perspective by examining the gradient propagation during distillation.
\item Based on the analysis, we propose the AEKT loss that can facilitate adaptive learning of explicit knowledge provided by the teacher model. 
\item We present a simple but effective technique to serialize the classification task and the distillation task for effective distillation of the teacher's knowledge.
\end{itemize}
\section{Related works}
\label{sec:related_works}

The classical KD method proposed in \cite{hinton14distilling} first demonstrates that the teacher model's knowledge can be transferred to a student model by minimizing the KL divergence between the probability distributions of the two models. 
It is often explained that the softened labels obtained from the teacher model, which contains implicit knowledge in the probabilities for the non-target classes, are valuable hints to train the student model \cite{hinton14distilling}.
Some studies also acknowledge a regularization effect by the softened labels \cite{yuan20rethinking,zhou21rethinking}.

Feature-based KD trains the student model in a way that the features from its intermediate layers become similar to those of the teacher model.
FitNet \cite{romero15fitnet} distills the intermediate features of the teacher to train a student model that is deeper and thinner than the teacher.
In the attention transfer (AT) method \cite{zagoruyko17paying}, the student is trained to mimic the attention map of the teacher.
The work in \cite{heo19comprehensive} proposes a method (abbreviated as OFD) by considering design aspects including teacher transform, student transform, distillation feature position, and distance function.
Relational KD (RKD) \cite{park19relational} transfers the relationship among the features of different data.
Contrastive representation distillation (CRD) \cite{tian20contrastive} employs a contrastive learning approach for distillation.
Softmax regression representation learning (SRRL) \cite{yang21knowledge} focuses on matching the features at the penultimate layers of the teacher and the student.
Knowledge review (KR) \cite{chen21distilling} proposes to transfer knowledge among different levels.

Logit-based KD tries to match the logits between the teacher and the student.
It is advantageous over feature-based KD in terms of efficiency and simplicity, but its task performance such as classification accuracy is often lower than that of feature-based KD.
To overcome this drawback, several methods have been proposed.
One such method is DIST \cite{huang22knowledge}, which replaces the KL divergence with the correlation coefficient to maintain inter-class and intra-class relations between the teacher and the student.
DKD \cite{zhao22decoupled} proposes to highlight the non-target class information in the loss function to transfer implicit knowledge effectively.
Normalized KD (NKD) \cite{yang23from} also considers effective transfer of implicit knowledge by normalizing the non-target class probabilities.
To distill between heterogeneous architectures without architecture-specific information, One-for-all (OFA) \cite{hao2023oneforall} projects intermediate features into logits.
Different from these methods, the proposed method focuses on effective transfer of explicit knowledge via adaptive learning of the target class information.
\section{Our approach}
\label{sec:Our_approach}

\subsection{Gradient analysis}

\paragraph{Classical KD and DKD.} 

We first briefly review the ideas of the classical KD \cite{hinton14distilling} and DKD \cite{zhao22decoupled}.
Let $\mathbf{p} = [p_1, ..., p_C]$ denote the output probability distribution, where $C$ is the number of classes.
Then, the loss of the classical KD is given by
\begin{equation}
\label{eq:kd}
	\mathcal{L}_{KD} = KL \left(\mathbf{p}^T \Vert \mathbf{p}^S \right)  = \sum_{i=1}^C p_i^T (\log p_i^T - \log p_i^S ),
\end{equation}
where the superscripts $T$ and $S$ denote the teacher and student models, respectively. Without loss of generality, we omit the temperature parameter.

In \cite{zhao22decoupled}, it is shown that Eq.~\ref{eq:kd} can be re-written as
\begin{align}
\label{eq:kd_recon}
	\mathcal{L}_{KD} = \underbrace{KL\left(\left[p_t^T,p_{\neg t}^T\right] \Vert \left[p_t^S,p_{\neg t}^S\right]\right)}_{\mathcal{L}_{TCKD}}  +\ p_{\neg t}^T \ \underbrace{KL\left( \hat{\mathbf{p}}^T \Vert \hat{\mathbf{p}}^S \right)}_{\mathcal{L}_{NCKD}},
\end{align}
where $p_t$ is the probability for the target class and $p_{\neg t} = \sum_{i=1, i\neq t}^C p_i$ is the sum of the probabilities for the non-target classes. And, $\hat{\mathbf{p}}\in\mathbb{R}^{C-1}$ is the re-normalized probability distribution for the non-target classes, i.e.,
$\hat{\mathbf{p}} = [\hat{p}_1, ..., \hat{p}_{t-1}, \hat{p}_{t+1}, ..., \hat{p}_{C} ]$
where $\hat{p}_{i} = p_i / p_{\neg t}$.
In other words, the loss of the classical KD can be divided into two terms: the loss for target class KD (TCKD), denoted as $\mathcal{L}_{TCKD}$, and the loss for non-target class KD (NCKD), denoted as $\mathcal{L}_{NCKD}$.
The former measures the similarity of the binary probabilities (target vs.~non-target) from the teacher and the student, while the latter compares the probability distributions only for the non-target classes, which plays a role to deliver the implicit knowledge of the teacher to the student.
It can be seen that the NCKD term is weighted by $p_{\neg t}^T=1-p_t^T$. 
This means that when the teacher's confidence for the target class is high, the student can hardly learn the implicit knowledge.
To avoid this issue, DKD uses a constant weight for the NCKD term:
\begin{align}
\label{eq:dkd_loss}
	\mathcal{L}_{DKD} = \alpha \mathcal{L}_{TCKD} + \beta \mathcal{L}_{NCKD}.
\end{align}

\paragraph{Gradients to student model.}

As a new perspective to analyze the learning processes of KD and DKD, we consider the gradients backpropagated to the student model's logits.
In other words, we examine the gradients of the loss ($\mathcal{L}_{KD}$ or $\mathcal{L}_{DKD}$) with respect to the logits of the student model. 
Let $z_i^S$ denote the $i$th class logit of the student model, from which the class probability is computed as the softmax output:
\begin{align}
	p_i^S = \frac{e^{z_i^S}}{\sum_{l=1}^C e^{z_l^S}}
\end{align}

First, let us examine the gradients in the classical KD (more details in Section \ref{sec:appendix_kd} of Appendix).
\begin{align}
	\label{eq:kd_partial}
	\frac{\partial\mathcal{L}_{KD}}{\partial z_i^S} &= \sum_{k=1}^C \frac{\partial\mathcal{L}_{KD}}{\partial p_k^S} \frac{\partial p_k^S}{\partial z_i^S} \ 
	= \frac{\partial\mathcal{L}_{KD}}{\partial p_i^S} \frac{\partial p_i^S}{\partial z_i^S} + \sum_{\substack{k=1\\k\neq i}}^C \frac{\partial\mathcal{L}_{KD}}{\partial p_k^S} \frac{\partial p_k^S}{\partial z_i^S} \ 
	= p_i^S - p_i^T.
\end{align}
Thus, The gradient backpropagated to the logit of the student model for a class is given by the probability difference for the class between the student and the teacher.

Next, we examine the gradients in DKD. 
According to Eq.~\ref{eq:dkd_loss}, we can separately obtain the gradients of the TCKD and NCKD loss terms, and then add them using weights $\alpha$ and $\beta$.
The TCKD loss term is written as
\begin{align}
\label{eq:tckd}
	\mathcal{L}_{TCKD} = p_t^T (\log p_t^T - \log p_t^S) + p_{\neg t}^T (\log p_{\neg t}^T - \log p_{\neg t}^S),
\end{align}
where 
\begin{align}
	p_{\neg t}^S = \frac{\sum_{m=1,m\neq t}^C e^{z_m^S}}{\sum_{l=1}^C e^{z_l^S}}.
\end{align}
The gradient of $\mathcal{L}_{TCKD}$ with respect to $z_t^S$ can be written as
\begin{align}
	\label{eq:grad_tckd1}
	\frac{\partial\mathcal{L}_{TCKD}}{\partial z_t^S} &= \frac{\partial\mathcal{L}_{TCKD}}{\partial p_t^S} \frac{\partial p_t^S}{\partial z_t^S} + \frac{\partial\mathcal{L}_{TCKD}}{\partial p_{\neg t}^S} \frac{\partial p_{\neg t}^S}{\partial z_t^S} = p_t^S - p_t^T.
\end{align}
The gradient of $\mathcal{L}_{TCKD}$ with respect to $z_i^S$ ($i \neq t$) is obtained as (more details in Section \ref{sec:appendix_dkd})
\begin{align}
	\label{eq:grad_tckd2}
	\frac{\partial\mathcal{L}_{TCKD}}{\partial z_i^S} &= \frac{\partial\mathcal{L}_{TCKD}}{\partial p_t^S} \frac{\partial p_t^S}{\partial z_i^S} + \frac{\partial\mathcal{L}_{TCKD}}{\partial p_{\neg t}^S} \frac{\partial p_{\neg t}^S}{\partial z_i^S} = \left( 1- \frac{p_{\neg t}^T}{p_{\neg t}^S} \right) p_i^S.
\end{align}

The NCKD loss term is written as
\begin{align}
\label{eq:nckd}
	\mathcal{L}_{NCKD} = \sum_{\substack{i=1\\i\neq t}}^C \hat{p}_i^T (\log \hat{p}_i^T - \log \hat{p}_i^S),
\end{align}
where 
\begin{align}
	\hat{p}_i^S = \frac{ e^{z_i^S}}{\sum_{l=1, l\neq t}^C e^{z_l^S}}.
\end{align}
The gradient of $\mathcal{L}_{NCKD}$ with respect to $z_i^S$ ($i \neq t$) is given by
\begin{align}
	\frac{\partial\mathcal{L}_{NCKD}}{\partial z_i^S} &=
\sum_{k=1}^C \frac{\partial\mathcal{L}_{NCKD}}{\partial \hat{p}_k^S} \frac{\partial \hat{p}_k^S}{\partial z_i^S},
\end{align}
which is equivalent to Eq.~\ref{eq:kd_partial} except that $p_k^S$ needs to be changed to $\hat{p}_k^S$.
Thus, 
\begin{align}
	\label{eq:grad_nckd}
	\frac{\partial\mathcal{L}_{NCKD}}{\partial z_i^S} = \hat{p}_i^S - \hat{p}_i^T = \frac{1}{p_{\neg t}^S} p_i^S - \frac{1}{p_{\neg t}^T} p_i^T.
\end{align}

By combining Eqs.~\ref{eq:dkd_loss}, \ref{eq:grad_tckd1}, \ref{eq:grad_tckd2}, and \ref{eq:grad_nckd}, the gradients for the target and non-target classes can be written as 
\begin{align}
	\frac{\partial\mathcal{L}_{DKD}}{\partial z_t^S} &= \alpha(p_t^S - p_t^T), \label{eq:dkd_partial_t}\\
	\frac{\partial\mathcal{L}_{DKD}}{\partial z_i^S} &= \left\{ \alpha \underline{\left( 1- \frac{p_{\neg t}^T}{p_{\neg t}^S} \right) } + \frac{\beta}{p_{\neg t}^S} \right\} p_i^S - \frac{\beta}{p_{\neg t}^T} p_i^T ~~ (i\neq t), \label{eq:dkd_partial_nott}
\end{align}
respectively.
For the target class, the gradient in DKD is the same to that in the classical KD (Eq.~\ref{eq:kd_partial}).
The difference of DKD and classical KD lies in the gradient for the non-target classes (Eq.~\ref{eq:dkd_partial_nott}).
The two terms containing $\beta$ are basically the same to Eq.~\ref{eq:kd_partial} but they are normalized by the respective probabilities for the non-target classes.
Thus, even when the teacher model is highly confident for the target class and thus $p_i^T$ is small, its implicit knowledge can be still delivered to the student through the gradient.
More prominently, the underlined first term in Eq.~\ref{eq:dkd_partial_nott} is newly introduced in DKD.
It adjusts the weight for $p_i^S$ in a way that when the non-target probability from the teacher is large, i.e., the teacher's implicit knowledge is rich, the weight for the student's probability $p_i^S$ is reduced so that the teacher's probability $p_i^T$ influences more to the gradient. 
In other words, unlike the classical KD, the gradients for the non-target logits in DKD, which allows learning of the teacher's implicit knowledge, are dynamically adjusted according to the probability values.

\subsection{Adaptive Explicit Knowledge Transfer (AEKT) loss}
\label{sec:proposed}

The previous analysis highlights the benefit of dynamic adjustment of the gradients for effective learning of implicit knowledge.
Then, we pay attention to the opportunity to apply a similar dynamic mechanism for learning of explicit knowledge.
For this, we propose a new loss called adaptive explicit knowledge transfer (AEKT) loss that adaptively weights the degree of learning about the target class based on the ratio of the teacher's and student's confidence levels for the target class.

The proposed AEKT loss is defined as
\begin{align}
\label{eq:loss_AEKT}	
	\mathcal{L}_{AEKT} = \log \left( \frac{p_t^T}{p_t^S} \right) \cdot \left( 1- 2^{1- \frac{p_t^T}{\reallywideacute{p}_t^S}} \right),
\end{align}
where $\reallywideacute{p}_t^S$ indicates that $p_t^S$ is considered as a constant for computing the gradients, which can be accomplished by cloning and excluding it from the computational graph (using \texttt{.clone().detach()} function in PyTorch).
The new loss is basically a function of the probability ratio $p_t^T / p_t^S$.
The second component, $1- 2^{1- \frac{p_t^T}{\reallywideacute{p}_t^S}}$, is designed in such a way that it ranges from -1 to 1 as $p_t^T/p_t^S$ changes from 0 (when $p_t^T=0$) to $\infty$ (when $p_t^S=0$) and additionally becomes 0 when $p_t^T=p_t^S$.
The total distillation loss in the proposed method is given by 
\begin{align}
\label{eq:total_loss}
	\mathcal{L}_{total} = \alpha\mathcal{L}_{TCKD} + \beta\mathcal{L}_{NCKD} + \gamma \mathcal{L}_{AEKT}.
\end{align}

The gradient of $\mathcal{L}_{AEKT}$ with respect to $z_t^S$ can be computed as follows (details in Section \ref{sec:appendix_aekt}).
\begin{align}
\label{eq:grad_aekt_t}
	\frac{\partial \mathcal{L}_{AEKT}}{\partial z_t^S} &= \frac{\partial \mathcal{L}_{AEKT}}{\partial p_t^S} \frac{\partial p_t^S}{\partial z_t^S} =\
        -(1-p_t^S) \left( 1- 2^{1- \frac{p_t^T}{\reallywideacute{p}_t^S}} \right).
\end{align}
Therefore, the gradient of $\mathcal{L}_{total}$ for the target class becomes
\begin{align}
	\frac{\partial\mathcal{L}_{total}}{\partial z_t^S} &= \alpha(p_t^S - p_t^T) \underline{ -\gamma  (1-p_t^S)  \left( 1- 2^{1- \frac{p_t^T}{\reallywideacute{p}_t^S}} \right) } , \label{eq:total_partial_t}
\end{align}
The underlined term is newly introduced by $\mathcal{L}_{AEKT}$, and it dynamically adjusts the gradient for the target class.
When the confidence ratio for the target class ($p_t^T/p_t^S$) is larger than 1, $1- 2^{1- \frac{p_t^T}{\reallywideacute{p}_t^S}}$ becomes positive and the underlined term has the same sign to $-\alpha p_t^T$, implying that the contribution of the teacher's information to the gradient becomes larger.
This process is controlled by the student's confidence $p_t^S$ (in $1-p_t^S$) in a way that as the student is less confident for the target class, the contribution of the teacher's information becomes even more larger. 
In contrast, when the student is more confident than the teacher so that $p_t^T/p_t^S<1$, the term $1- 2^{1- \frac{p_t^T}{\reallywideacute{p}_t^S}}$ becomes negative, and the gradient is pulled away from the teacher's information.
When the student is highly confident, $z_t^S$ is already saturated and thus the amount of the gradient adjustment does not have to be large, which is reflected in the term $1-p_t^S$.

And, the gradient of $\mathcal{L}_{AEKT}$ with respect to $z_i^S$ ($i\neq t$) is written as (details in Section \ref{sec:appendix_aekt})
\begin{align}
	\frac{\partial \mathcal{L}_{AEKT}}{\partial z_i^S} &=
    \frac{\partial \mathcal{L}_{AEKT}}{\partial p_t^S} \frac{\partial p_t^S}{\partial z_i^S} = -\frac{1}{p_t^S} \left( 1- 2^{1- \frac{p_t^T}{\reallywideacute{p}_t^S}} \right) (-p_t^S p_i^S ) = \left( 1- 2^{1- \frac{p_t^T}{\reallywideacute{p}_t^S}} \right) p_i^S.
    \label{eq:grad_aekt_nt}
\end{align}

Thus, the gradient of $\mathcal{L}_{total}$ for a non-target class $i (\neq t)$ is 
 \begin{align}
	\frac{\partial\mathcal{L}_{total}}{\partial z_i^S} = &\left\{ \gamma \underline{ \left( 1- 2^{1- \frac{p_t^T}{\reallywideacute{p}_t^S}} \right) } + \alpha \left( 1- \frac{p_{\neg t}^T}{p_{\neg t}^S} \right)  + \frac{\beta}{p_{\neg t}^S} \right\} p_i^S - \frac{\beta}{p_{\neg t}^T} p_i^T.
	\label{eq:total_partial_nott}
\end{align}
The underlined term stems from $\mathcal{L}_{AEKT}$.
It can be easily verified that the effect of this term is aligned with the following term (i.e., $\alpha \left( 1- \frac{p_{\neg t}^T}{p_{\neg t}^S} \right)$).
Thus, it does not impose an issue in learning implicit knowledge.

\subsection{Task serialization}\label{Task Serialization}

\begin{wrapfigure}{r}{0.4\textwidth}
    \vspace{-5pt}
    \begin{center}
        \includegraphics[width=0.35\textwidth]{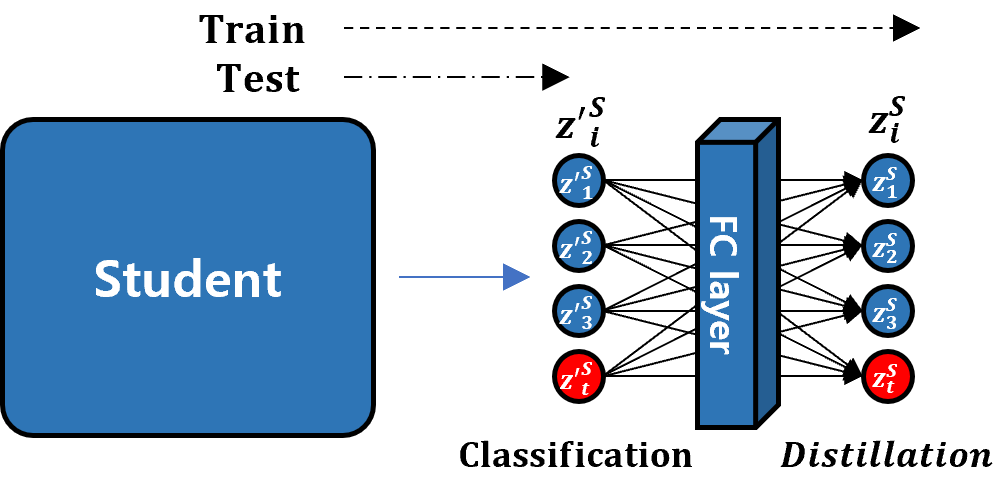}
    \end{center}
    \vspace{-10pt}
    \caption{\textbf{Task serialization with an additional FC layer.}}
    \label{fig:serial} 
    \vspace{-10pt}
\end{wrapfigure}

In logit-based KD, it is usual to employ the classification loss (i.e., cross-entropy (CE)) along with the distillation loss, which are applied on the same level, i.e., the distillation loss at the logits and the CE loss at the softmax outputs from the logits.
This means that the logits of the student model need to be trained to satisfy two different probability distributions: one having a probability of one for the target class and zero for all other classes, and the other representing the softened distribution extracted from the teacher model.
We argue that enforcing the two different tasks (i.e., classification and distillation) simultaneously could limit effective learning of the student model.

To resolve this issue, we propose to serialize the two tasks.
As shown in \autoref{fig:serial}, a linear FC layer is added to take the logits of the student model as input.
It then outputs a transformed set of logits, for which the distillation task is performed as described in Section~\ref{sec:proposed}.
Note that the added FC layer is used only during training; the logits of the student (before the added FC layer) are used for inference.

\begin{figure*}[t]
    \centerline{\includegraphics[width=.9\textwidth]{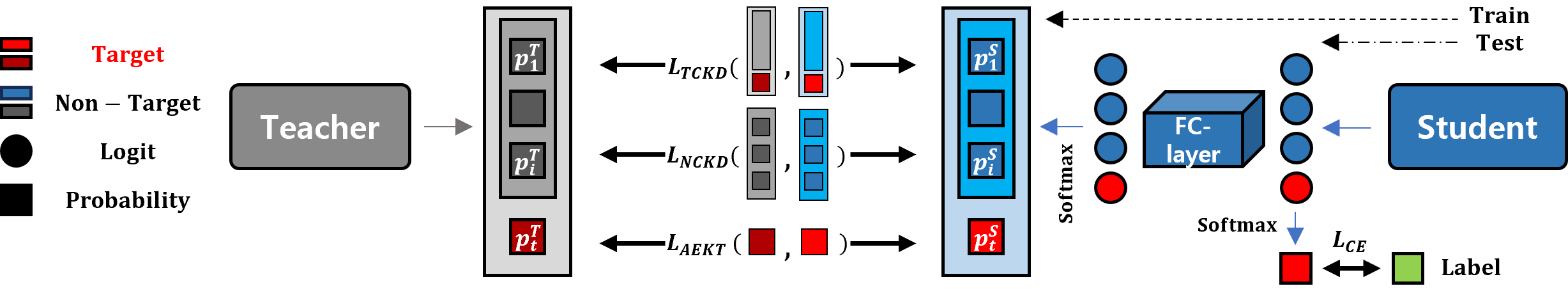}}
    \caption{\textbf{Overall architecture of the proposed AEKT method.} Three losses for distillation, $\mathcal{L}_{TCKD}$, $\mathcal{L}_{NCKD}$, and $\mathcal{L}_{AEKT}$, are measured between the output probabilities of the teacher ($p_i^T$) and those of the student ($p_i^S$). A FC layer is used to serialize the distillation task and the classification task. The cross-entropy (CE) loss is measured before the FC layer. The FC layer is used only for training of the student; the inference is performed using the logits of the student model.}
    \label{fig:overall} 
    \vspace{-10pt}
\end{figure*}

Let ${z'}_i^S$ denote the logit for the $i$th class from the student (before transformation by the added FC layer), and $w_{ij}\in\mathbb{R}^{C\times C}$ the weight parameter from ${z'}_i^S$ to ${z}_j^S$ in the FC layer.
The gradients backpropagated to the logits of the student model are written as
\begin{align}
	\frac{\partial \mathcal{L}_{total}}{\partial {z'}_i^S} = \sum_{j=1}^C w_{ij} \frac{\partial \mathcal{L}_{total}}{\partial {z}_j^S}.
	\label{eq:ser}
\end{align}
Without the FC layer, the gradient propagated to a student's logit (Eqs.~\ref{eq:total_partial_t} and \ref{eq:total_partial_nott}) is computed only using the information of the corresponding class:
For the target class (Eq.~\ref{eq:total_partial_t}), the gradient uses solely the target class probabilities of the teacher and the student.
For a non-target class (Eq.~\ref{eq:total_partial_nott}), only the summed probabilities for the non-target classes (but not individually) are involved along with the probabilities of the class.
As can be seen in Eq.~\ref{eq:ser}, however, the added FC layer enables to consider the relationship among classes during training.

The overall architecture of the proposed method combining the new loss and task serialization is illustrated in \autoref{fig:overall}.
\section{Experiments}
\label{sec:Experiments}

We evaluate the proposed method on the CIFAR-100 \cite{cifar100} and ImageNet \cite{imagenet} datasets in comparison to state-of-the-art feature-based KD methods (FitNet \cite{romero15fitnet}, AT \cite{zagoruyko17paying}, OFD \cite{heo19comprehensive}, RKD \cite{park19relational}, CRD \cite{tian20contrastive}, SRRL \cite{yang21knowledge}, and KR \cite{chen21distilling}) and logit-based KD methods (classical KD \cite{hinton14distilling}, DKD \cite{zhao22decoupled}, DIST \cite{huang22knowledge}, NKD \cite{yang23from}, and OFA \cite{hao2023oneforall}). 
Further analysis results are also presented.
All detailed experimental settings can be found in Section \ref{sec:appendix_experimental_setting} of Appendix.

\subsection{CIFAR-100}


By following \cite{zhao22decoupled}, we compare the performance of different methods in two scenarios:
The first scenario uses the same architecture for both the teacher and the student, including ResNet \cite{resnet}, WRN \cite{wrn}, and VGG \cite{vgg}. 
In the second scenario, the teacher and the student have different architectures; in particular, the student is a model designed for improved efficiency, including ShuffleNet \cite{shufflenetv1,shufflenetv2} and MobileNet \cite{mobilenetv1,mobilenetv2}, while the teacher is ResNet \cite{resnet}, WRN \cite{wrn}, or VGG \cite{vgg}.
The pairs of the teacher and student models are set as in \cite{zhao22decoupled}, and we use the trained teacher models provided by \cite{zhao22decoupled}. 




\autoref{tab:CIFAR-100_SAME} and \autoref{tab:CIFAR-100_DIFF} show the comparison results on CIFAR-100. 
In most cases, our method achieves the best or second-best performance, which shows that the proposed method is effective.

\begin{table*}[!t]
    \setlength{\tabcolsep}{8pt}        
    \renewcommand{\arraystretch}{1.2}   
    \centering
    \resizebox{0.85\textwidth}{!}
    {
    \begin{tabular}{c|c|cccccc}
        \Xhline{2pt}
        \multirow{2}{*}{Model}&Teacher&ResNet56&ResNet110&ResNet32x4&WRN-40-2&WRN-40-2&VGG13 \\
        \cline{2-8}
        & Student&ResNet20&ResNet32&ResNet8x4&WRN-16-2&WRN-40-1&VGG8 \\
        \Xcline{1-8}{2pt}
        \multirow{2}{*}{Baseline}&Teacher&72.34&74.31&79.42&75.61&75.61&74.64 \\
        \cline{2-8}
        & Student&69.06&71.14&72.50&73.26&71.98&70.36 \\
        \Xcline{1-8}{2pt}
        \multirowcell{6}{Feature-\\based}%
        & FitNet \cite{romero15fitnet}   & 69.21 & 71.06 & 73.50 & 73.58 & 72.24 & 71.02 \\
        & OFD \cite{heo19comprehensive}  & 70.98 & 73.23 & 74.95 & 75.24 & 74.33 & 73.95 \\
        & RKD \cite{park19relational}    & 69.61 & 71.82 & 71.90 & 73.35 & 72.22 & 71.48 \\
        & CRD \cite{tian20contrastive}   & 71.16 & 73.48 & 75.51 & 75.48 & 74.14 & 73.94 \\
        & SRRL \cite{yang21knowledge}    & 71.44 & 73.80 & 75.92 & 75.96 & 74.75 & 74.40 \\
        & KR \cite{chen21distilling}     & 71.89 & 73.89 & 75.63 & 76.12 & \underline{75.09} & 74.84 \\
        \Xcline{1-8}{2pt}
        \multirowcell{6}{Logit-\\based}%
        & KD \cite{hinton14distilling}   & 70.66 & 73.08 & 73.33 & 74.92 & 73.54 & 72.98 \\
        & DKD \cite{zhao22decoupled}     & 71.97 & 74.11 & 76.32 & 76.24 & 74.81 & 74.68 \\
        & DIST \cite{huang22knowledge}   & 71.75 & 73.91*& 76.31 & 75.63*& 74.73 & 73.96*\\
        & NKD \cite{yang23from}          & 71.18*& 73.50*& 76.35 & 75.43*& 74.25*& \underline{74.86} \\
        \cline{2-8}
        & \textbf{AEKT (w/o Ser.)}       & \underline{72.05} & \underline{74.16} & \underline{76.81} & \textbf{76.43} & 74.97 & \textbf{75.03} \\
        & \textbf{AEKT}                  & \textbf{72.15} & \textbf{74.29} & \textbf{77.03} & \underline{76.34} & \textbf{75.11} & 74.79 \\
        \Xhline{2pt}
    \end{tabular}
    }
    \caption{\textbf{Performance comparison in accuracy (\%) on the CIFAR-100 validation set for same architecture pairs of teachers and students.} The results of our method without task serialization is also shown. The values marked with `*' are from our own experiments, which were not provided by the respective original papers. The best and second-best cases in each teacher-student pair are marked with bold faces and underlines, respectively.}
    \label{tab:CIFAR-100_SAME}
\end{table*}
\begin{table*}[t!]
    \vspace{-5pt}
    \setlength{\tabcolsep}{8pt}        
    \renewcommand{\arraystretch}{1.2}   
    \centering
    \resizebox{0.85\textwidth}{!}
    {
    \begin{tabular}{c|c|ccccc}
        \Xhline{2pt}
        \multirow{2}{*}{Model}&Teacher&ResNet32x4&WRN-40-2&VGG13&ResNet50&ResNet32x4 \\
        \cline{2-7}
        & Student&ShuffleNet-V1&ShuffleNet-V1&MobileNet-V2&MobileNet-V2&ShuffleNet-V2 \\
        \Xcline{1-7}{2pt}
        \multirow{2}{*}{Baseline}&Teacher&79.42&75.61&74.64&79.34&79.42 \\
        \cline{2-7}
        & Student&70.50&70.50&64.60&64.60&71.82 \\
        \Xcline{1-7}{2pt}
        \multirowcell{6}{Feature-\\based}%
        & FitNet \cite{romero15fitnet}   & 73.59 & 73.73 & 64.14 & 63.16 & 73.54  \\
        & OFD \cite{heo19comprehensive}  & 75.98 & 75.85 & 69.48 & 69.04 & 76.82 \\
        & RKD \cite{park19relational}    & 72.28 & 72.21 & 64.52 & 64.43 & 73.21 \\
        & CRD \cite{tian20contrastive}   & 75.11 & 76.05 & 69.73 & 69.11 & 75.65 \\
        & SRRL \cite{yang21knowledge}    & 75.66 & 76.61 & 69.14 & 69.45 & 76.40 \\
        & KR \cite{chen21distilling}     & \textbf{77.45} & \underline{77.14} & 70.37 & 69.89 & \textbf{77.78} \\
        \Xcline{1-7}{2pt}
        \multirowcell{6}{Logit-\\based}%
        & KD \cite{hinton14distilling}   & 74.07 & 74.83 & 67.37 & 67.35 & 74.45 \\
        & DKD \cite{zhao22decoupled}     & 76.45 & 76.70 & 69.71 & 70.35 & 77.07 \\
        & DIST \cite{huang22knowledge}   & 76.34 & 76.19*& 68.21*& 68.66 & 77.35 \\
        & NKD \cite{yang23from}          & 76.31*& 76.46*& 70.22 & 70.67 & 76.92* \\
        \cline{2-7}
        & \textbf{AEKT (w/o Ser.)}       & 76.73 & 77.05 & \underline{70.39} & \underline{70.89} & 77.18\\
        & \textbf{AEKT}                  & \underline{77.03} & \textbf{77.21} & \textbf{70.41} & \textbf{71.09} & \underline{77.39} \\
        \Xhline{2pt}
    \end{tabular}
    }
    \caption{\textbf{Performance comparison in accuracy (\%) on the CIFAR-100 validation set for different architecture pairs of teachers and students.} The results of our method without task serialization is also shown. The values marked with `*' are from our own experiments, which were not provided by the respective original papers. The best and second-best cases in each teacher-student pair are marked with bold faces and underlines, respectively.}
    \label{tab:CIFAR-100_DIFF}
    \vspace{-5pt}
\end{table*}

\begin{table}[t]
    \centering
    \resizebox{\textwidth}{!}{
    \begin{tabular}{cc|cc|ccccc|ccc|cc}
        \Xhline{2pt}
        \multicolumn{2}{c|}{Model}&\multicolumn{2}{c|}{Baseline}&\multicolumn{5}{c|}{Feature-based}&\multicolumn{5}{c}{Logit-based}\\
        \Xhline{2pt}
        Teacher&Student&Teacher&Student& OFD\cite{heo19comprehensive} & RKD\cite{park19relational} & CRD\cite{tian20contrastive} & SRRL\cite{yang21knowledge} & KR\cite{chen21distilling} & KD\cite{hinton14distilling}& DKD\cite{zhao22decoupled} & DIST\cite{huang22knowledge}& \textbf{AEKT (w/o Ser.)}& \textbf{AEKT}\\
        \hline
        ResNet34 & ResNet18 & 73.31 & 69.75 & 70.81 & 71.34 & 71.17 & 71.73 & 71.61 & 70.66 & 71.70 & \textbf{72.07} & 71.75 & \underline{71.86} \\
        ResNet50 & MobileNet-V1 & 76.16 & 68.87 & 71.25 & 71.32 & 71.37 & 72.49 & 72.56 & 68.58 & 72.05 & \underline{73.24} & 73.18 & \textbf{73.30} \\
        \Xhline{2pt}
    \end{tabular}}
    \caption{\textbf{Performance comparison in accuracy (\%) on the ImageNet validation set with CNN architectures.} The results of our method without task serialization is also shown. The best and second-best cases in each teacher-student pair are marked with bold faces and underlines, respectively.}
    \label{tab:IMAGENET}
\end{table}
\begin{table}[t]
    \vspace{-10pt}
    \centering
    \resizebox{\textwidth}{!}{
    \begin{tabular}{cc|cc|ccc|cccc|cc}
        \Xhline{2pt}
        \multicolumn{2}{c|}{Model}&\multicolumn{2}{c|}{Baseline}&\multicolumn{3}{c|}{Feature based}&\multicolumn{6}{c}{Logit based}  \\
        \Xhline{2pt}
        Teacher&Student&Teacher&Student&FitNet\cite{romero15fitnet}&RKD\cite{park19relational}&CRD\cite{tian20contrastive}&KD\cite{hinton14distilling}&DKD\cite{zhao22decoupled}&DIST\cite{huang22knowledge}&OFA\cite{hao2023oneforall}&AEKT (w/o Ser.)&AEKT \\ 
        \hline
        DeiT-T & ResNet18 & 72.17 & 69.75 & 70.44 & 69.47 & 69.25 & 70.22 & 69.39 & 70.64 & \textbf{71.34} & 70.77 & \underline{71.20} \\
        DeiT-T & MobileNet-V1 & 72.17 & 68.87 & 70.95 & 69.72 & 69.60 & 70.87 & 70.14 & 71.08 & 71.39 & \underline{71.51} & \textbf{72.25} \\
        Swin-T & ResNet18 & 81.38 & 69.75 & 71.18 & 68.89 & 69.09 & 71.14 & 71.10 & 70.91 & \textbf{71.85} & 71.02 & \underline{71.43} \\
        Swin-T & MobileNet-V1 & 81.38 & 68.87 & 71.75 & 67.52 & 69.58 & 72.05 & 71.71 & 71.76 & 72.32 & \textbf{72.89} & \underline{72.62} \\
        \Xhline{2pt}
    \end{tabular}}
    \caption{\textbf{Performance comparison in accuracy (\%) on the ImgeNet validation set with Transformer architectures.} The results of our method without task serialization is also shown. The best and second-best cases in each teacher-student pair are marked with bold faces and underlines, respectively.}
    \label{tab:IMAGENET_TRANSFORMER}
    \vspace{-3pt}
\end{table}

\subsection{ImageNet}
\label{sec:Experiments_ImageNet}
As in \cite{zhao22decoupled}, we evaluate our method on the ImageNet dataset with CNN architectures, i.e., ResNet-34$\rightarrow$ResNet-18 and ResNet-50$\rightarrow$MobileNet-V1.
We also consider the cases with Transformer architectures, i.e., DeiT-T \cite{deit} or Swin-T \cite{swin} as the teacher and ResNet-18 or MobileNet-V1 as the student.

\autoref{tab:IMAGENET} and \autoref{tab:IMAGENET_TRANSFORMER} summarize the performance comparison results. 
In \autoref{tab:IMAGENET}, the proposed AEKT method is comparable to the best existing method (i.e., DIST \cite{huang22knowledge}), showing the highest or second highest accuracy.
We also found that our method can be improved further via hyperparameter tuning up to 72.20\% for ResNet34$\rightarrow$ResNet18, which is the best performance.
\begin{table}[h!]
    \vspace{-10pt}
    \centering
    \resizebox{\textwidth}{!}{
    \begin{tabular}{c|cccccc|cccc|c}
         \Xhline{2pt}
         Metric & FitNet\cite{romero15fitnet} & OFD \cite{heo19comprehensive} & RKD \cite{park19relational} & CRD \cite{tian20contrastive} & SRRL \cite{yang21knowledge} & KR \cite{chen21distilling} & KD \cite{hinton14distilling} & DKD \cite{zhao22decoupled} & DIST \cite{huang22knowledge} & NKD \cite{yang23from} & \textbf{AEKT} \\
         \Xhline{2pt}
         Cosine & 0.811 & 0.828 & 0.797 & 0.833 & 0.832 & 0.838 & 0.812 & 0.844 & 0.828 & \underline{0.847} & \textbf{0.848} \\
         \hline
         Agreement & 0.755 & 0.766 & 0.733 & 0.767 & 0.769 & 0.774 & 0.754 & \underline{0.787} & 0.778 & 0.782 & \textbf{0.790}\\
         \Xhline{2pt}
    \end{tabular}}
    \caption{\textbf{Student-teacher similarity.} The similarity is measured in terms of average cosine similarity and prediction agreement. The results are for ResNet32x4$\rightarrow$ResNet8x4 on CIFAR-100.}
    \label{tab:sim_val}
    \vspace{-15pt}
\end{table}
In \autoref{tab:IMAGENET_TRANSFORMER}, our method consistently performs well by showing the best or second-best performance for all cases. Note that OFA \cite{hao2023oneforall} is specifically designed for distillation between heterogeneous architectures, but it is outperformed by our method in an overall sense.
These results demonstrate that the proposed method is also effective on the large-scale classification problem.

\subsection{Analysis}

\paragraph{Correlation between teacher and student.}

The goal of KD is to make the student mimic the knowledge of the teacher. 
If the student has thoroughly learned the teacher's knowledge, the inference results of the teacher and the student would become similar.
To quantify the similarity between the teacher and student models, we employ two measures: cosine similarity and prediction agreement.
The cosine similarity is calculated between the output probability distributions of the teacher and the student.
The prediction agreement means the ratio of the data for which the teacher and the student yield the same classification results.
The results for ResNet32x4$\rightarrow$ResNet8x4 on CIFAR-100 are shown in \autoref{tab:sim_val}.
The proposed method shows the highest similarity between the teacher and the student, indicating that the goal of KD is achieved more successfully than the other methods.

\begin{wraptable}{r}{0.45\textwidth}
    \vspace{-10pt}
    \resizebox{0.45\textwidth}{!}{
    \begin{tabular}{c|c|cccccc|c}
        \Xhline{2pt}
                &All&(1)&(2)&(3)&(4)&(5)&(6)&Base \\
        \Xhline{2pt}
        $\alpha$& \checkmark & \checkmark & \checkmark &            & \checkmark &            &            & \\
        $\beta$ & \checkmark & \checkmark &            & \checkmark &            & \checkmark &            & \\
        $\gamma$& \checkmark &            & \checkmark & \checkmark &            &            & \checkmark & \\
        \hline
        AEKT(w/o Ser.)  & \textbf{76.81} & 76.32 & 66.61 & 76.31 & 68.81 & 75.20 & 67.97 & 72.50 \\
        AEKT            & \textbf{77.03} & 76.42 & 71.74 & 76.44 & 71.92 & 75.97 & 72.13 & 72.50 \\
        \Xhline{2pt}
    \end{tabular}}
    \caption{\textbf{Ablation study on the hyperparameters in our loss function on ResNet32x4→ResNet8x4.} `Base' means the case where the model is trained only with the conventional cross-entropy loss without KD.}
    \label{tab:Ablation_Loss}
    \vspace{-12pt}
\end{wraptable}

\paragraph{Ablation study on loss terms.}
To analyze the contribution of each term in our loss (TCKD, NCKD, and AEKT), we conduct additional experiments for CIFAR-100 with excluding one or two terms from our loss (by setting the corresponding weight to zero), and the results are summarized in \autoref{tab:Ablation_Loss}. We can make the following observations: 1) From (2), (4), and (6), we can see that $\beta$ affects performance gain
the most. When $\beta$ is zero, the student model’s performance with KD becomes worse than the base student model, which shows the importance of ‘implicit knowledge’ in KD. 2) From (1), (3), and (5), we can confirm that $\alpha$ and $\gamma$ have similar impact on performance gain. 3) All terms need to be used together in order to achieve the maximal performance. In other words, the three terms are complementary for the model performance.

\paragraph{Effect of AEKT loss.}
We additionally evaluate the effectiveness of our AEKT loss when it is applied to existing methods.
We choose KD \cite{hinton14distilling}, NKD \cite{yang23from}, and DIST \cite{huang22knowledge}, in which the AEKT loss can be plugged directly.
As shown in \autoref{tab:aekt_existing}, the proposed loss can boost the performance of KD and NKD by facilitating effective learning of explicit knowledge. 
In the case of DIST, we observe performance deterioration, probably because DIST employs the Pearson correlation coefficient instead of the KL divergence, while our loss is designed by considering the KL divergence.

\begin{figure}
    \begin{minipage}[c]{0.48\linewidth}
        \begin{subfigure}{0.48\textwidth}
            \includegraphics[width=\textwidth]{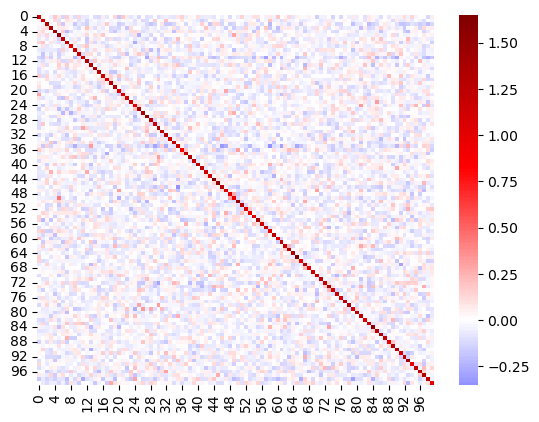}
            \caption{{ResNet32x4$\rightarrow$ResNet8x4 \\on CIFAR-100}}
        \end{subfigure}
        \hfill
        \begin{subfigure}{0.47\textwidth}
            \includegraphics[width=\textwidth]{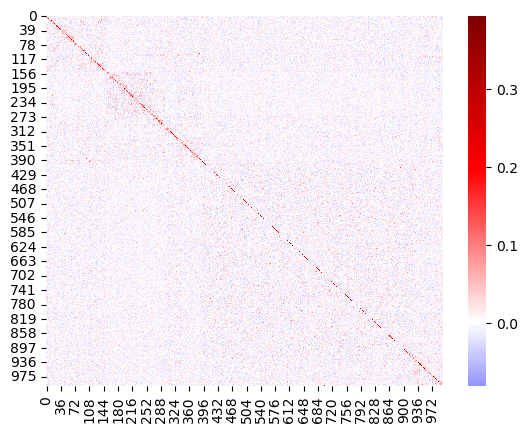}
            \caption{{ResNet34$\rightarrow$ResNet18 \\on ImageNet}}
        \end{subfigure}
        \caption{\textbf{Visualization of the learned weights of the FC layer for task serialization.} $w_{ij}$ is shown at the intersection the $i$th row and the $j$th column.}
        \label{fig:FC_Weight}
    \end{minipage}
    \hfill
    \begin{minipage}[c]{0.5\linewidth}
        \begin{subfigure}{0.48\textwidth}
            \includegraphics[width=\textwidth]{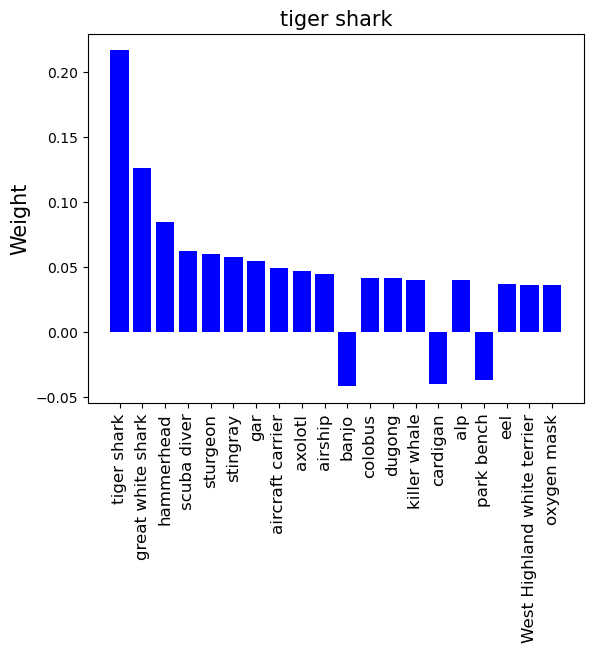}
        \end{subfigure}
        \hfill
        \begin{subfigure}{0.48\textwidth}
            \includegraphics[width=\textwidth]{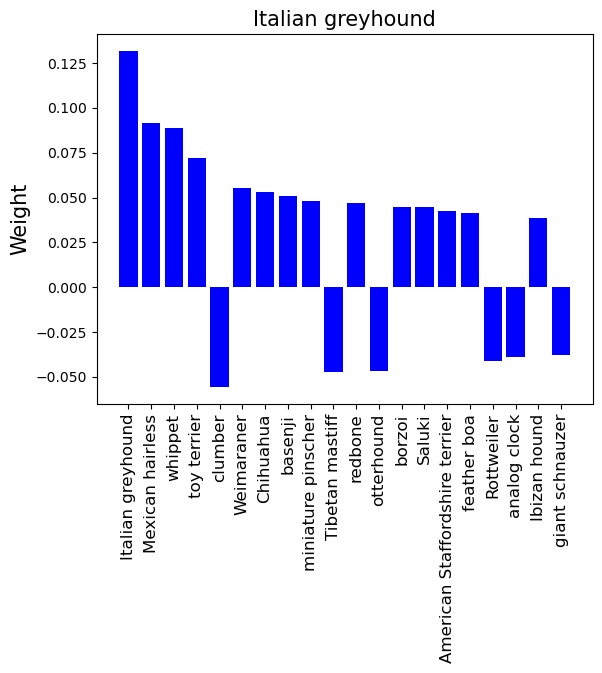}
        \end{subfigure}
        \caption{\textbf{Examples of the FC layer weights for ImageNet.} The 20 classes showing the largest absolute weight values are shown. Left: `tiger shark' class. Right: `Italian greyhound' class.}
        \label{fig:FC_weight_class}
    \end{minipage}
    \vspace{-10pt}
\end{figure}

\paragraph{Analysis of FC layer for serialization.}
To analyze the effect of task serialization by the added FC layer in our method, \autoref{fig:FC_Weight} visualizes the learned weights of the FC layer.
\begin{minipage}{.48\textwidth}
\centering
    \resizebox{\textwidth}{!}{
    \begin{tabular}{c|c|c|c}
        \Xhline{2pt}
        Method & KD \cite{hinton14distilling}  & NKD \cite{yang23from} & DIST \cite{huang22knowledge}  \\
        \Xhline{2pt}
        Acc. (+$\Delta$) & 74.70 (+0.48) & 76.37 (+0.19) & 76.22 (-0.12) \\
        \Xhline{2pt}
    \end{tabular}
    }
    \captionof{table}{\textbf{Performance when the proposed AEKT loss is plugged in existing methods.} The accuracy (\%) and the accuracy improvement ($\Delta$) on CIFAR-100 for ResNet32x4$\rightarrow$ResNet8x4 are shown.}
    \label{tab:aekt_existing}    
\end{minipage}
\hfill
\begin{minipage}{.48\textwidth}
\centering
    \resizebox{\textwidth}{!}{
    \begin{tabular}{c|c|c|c|c}
        \Xhline{2pt}
        Activation & $C$-$C$ & $C$-$C$-$C$ & $C$-$2C$-$C$ & $C$-$C/2$-$C$ \\
        \Xhline{2pt}
        Linear& \textbf{77.03} & 76.40 & 76.19 & 76.27   \\
        \cline{1-5}
        Nonlinear & 76.42 & 76.06 & 75.98 & 75.36   \\
        \Xhline{2pt}
    \end{tabular}
    }
    \captionof{table}{\textbf{Performance of different FC networks for serialization.} $C$-$K$-$C$ refers to the network structure having $C$ inputs, $K$ intermediate nodes, and $C$ outputs. The case of $C$-$C$ is the one used in the proposed method.}
    \label{tab:Ablation_Serial}
    \vspace{2pt}
\end{minipage}
The diagonal components ($w_{ii}$) have high values, indicating that the two logits ($z_i^S$ and ${z'}_i^S$) are highly correlated, and the logit for classification (${z'}_i^S$) contributes most to the logit of the same class for distillation (${z}_i^S$).
Nevertheless, the FC layer does not merely implement an identity matrix and many off-diagonal components are non-zero. 
This means that for the logit of a class, the information for other classes are exploited through the FC layer.
Some of the diagonal components for ImageNet have relatively small values, because the information of many related classes are involved in determining the logit of a class for distillation.
Interestingly, non-negligible portions of the weights are negative, which can be interpreted that the information of certain classes are used in a \textit{discriminative} way.
\autoref{fig:FC_weight_class} shows two examples of the top-20 weights for a class in a descending order of the absolute weight values.
The target class itself (`tiger shark' or `Italian greyhound') and classes with high semantic similarity (different types of sea creatures or dogs with similar appearances) are associated with large weight values.
On the left panel, the classes that are not related to the target class (`banjo', `cardigan', etc.) take negative weights.
On the right panel, negative weights are learned for some classes of dogs having dissimilar appearances (`Clumber', `Tibetan mastiff', etc.) and unrelated classes (`analog clock'). 
These observations confirm that task serialization not only eases simultaneous learning of the classification and distillation tasks but also allows to exploit inter-class relationship flexibly.

\paragraph{Networks for serialization.}
In our method, we use a linear FC layer having $C$ inputs and $C$ outputs for task serialization.
We conduct an experiment to evaluate different choices of the network for serialization.
Two FC layers having $C$, $2C$, or $C/2$ intermediate nodes are tested.
For each structure, we test the cases without and with a nonlinear activation function in each layer, i.e., PReLU (with a slope of 0.25 in the negative range) to facilitate learning of both positive and negative contributions shown in \autoref{fig:FC_Weight} and \autoref{fig:FC_weight_class}.
The results for ResNet32x4$\rightarrow$ResNet8x4 on CIFAR-100 are shown in \autoref{tab:Ablation_Serial}.
It is observed that an additional layer and use of the nonlinear activation function are not beneficial, and a linear FC layer is sufficient.

\section{Conclusion}
\label{sec:Conclusion}

We have proposed a new distillation method called AEKT, which includes a loss function promoting effective learning of explicit knowledge along with implicit knowledge, and a technique for serialization of the classification and distillation tasks. 
The effectiveness of the proposed method was shown through comparison to the state-of-the-art logit-based and feature-based KD methods.
Additional analysis results also supported the benefits of the proposed loss and the serialization technique.


{
    \small
    \bibliographystyle{ieeenat_fullname}
    \bibliography{AEKT}
}


\newpage
\section*{Appendix}
\appendix

\numberwithin{equation}{section}    
\numberwithin{figure}{section}
\numberwithin{table}{section}

\section{Derivation of the gradients}
\label{sec:appendix_derivation}
We provide more detailed derivations of the gradients shown in Section 3 of the main paper.

Let us first define the probabilities obtained from logits.
The probability for the $i$th class from the student, $p_i^S~(i=1,....,C)$, is written as
\begin{align}
	p_i^S = \frac{e^{z_i^S}}{\sum_{l=1}^C e^{z_l^S}}.
\end{align}
In particular, the probability for the target class from the student, $p_t^S$, can be written as
\begin{align}
	p_t^S = \frac{e^{z_t^S}}{\sum_{l=1}^C e^{z_l^S}}.
\end{align}

The sum of the probabilities for all non-target classes from the student, $p_{\neg t}^S$, is
\begin{align}
	p_{\neg t}^S = \frac{\sum_{m=1, m\neq t}^C e^{z_m^S}}{\sum_{l=1}^C e^{z_l^S}}.
\end{align}

The re-normalized probability for non-target classes, $\hat{p}_i^S=p_i^S/p_{\neg t}^S$ with $i\neq t$, is written as
\begin{align}
    \hat{p}_i^S = \frac{ e^{z_i^S}}{\sum_{l=1, l\neq t}^C e^{z_l^S}}.
\end{align}

\subsection{Classical KD (Eq.~\ref{eq:kd_partial})}
\label{sec:appendix_kd}
The classical KD loss using the KL divergence, $\mathcal{L}_{KD}$, is given by Eq.~1 of the main paper, i.e.,
\begin{align}
\mathcal{L}_{KD}= \sum_{i=1}^C p_i^T (\log p_i^T - \log p_i^S ).
\label{eq:loss_kd}
\end{align}
Using the chain rule, the gradient of $\mathcal{L}_{KD}$ with respect to the logit of the student $z_i^S$ can be written as
\begin{align}
    \frac{\partial\mathcal{L}_{KD}}{\partial z_i^S} &= \sum_{k=1}^C \frac{\partial\mathcal{L}_{KD}}{\partial p_k^S} \frac{\partial p_k^S}{\partial z_i^S}.
    \label{eq:grad_kd}
\end{align}

The first term is obtained by differentiating Eq.~\ref{eq:loss_kd} with respect to $p_k^S$:
\begin{align}
    \dfrac{\partial\mathcal{L}_{KD}}{\partial p_k^S}&=-\dfrac{p_k^T}{p_k^S}.
    \label{eq:kd_pk}
\end{align}

For the second term, there are two cases: $k=i$ and $k\neq i$.
When $k=i$,
\begin{align}
    \dfrac{\partial p_i^S}{\partial z_i^S} &=\dfrac{e^{z_i^S}\times\left(\sum_{l=1}^C e^{z_l^S}\right) -\left(e^{z_i^S}\right)^2}{\left(\sum_{l=1}^C e^{z_l^S}\right)^2} \nonumber \\ 
    &= \frac{e^{z_i^S}}{\sum_{l=1}^C e^{z_l^S}} - \left(\frac{e^{z_i^S}}{\sum_{l=1}^C e^{z_l^S}} \right)^2 \nonumber \\
    &= p_i^S-(p_i^S)^2 .
    \label{eq:kd_pi_zi}
\end{align}
Otherwise (i.e., $k\neq i$),
\begin{align}
    \dfrac{\partial p_k^S}{\partial z_i^S}
    &=\dfrac{-e^{z_k^S}\times e^{z_i^S}}{\left(\sum_{l=1}^C e^{z_l^S}\right)^2} \nonumber \\ 
    &=-\left(\dfrac{e^{z_k^S}}{\sum_{l=1}^C e^{z_l^S}} \right)
    \left(\dfrac{e^{z_i^S}}{\sum_{l=1}^C e^{z_l^S}} \right) \nonumber \\
    &= -p_k^S\cdot p_i^S \label{eq:kd_pk_zi}
\end{align}

Using Eqs.~\ref{eq:kd_pk}, \ref{eq:kd_pi_zi}, and \ref{eq:kd_pk_zi}, therefore, Eq.~\ref{eq:grad_kd} can be written as
\begin{align}
    \frac{\partial\mathcal{L}_{KD}}{\partial z_i^S} &= \sum_{k=1}^C \frac{\partial\mathcal{L}_{KD}}{\partial p_k^S} \frac{\partial p_k^S}{\partial z_i^S} \nonumber \\
    &= \frac{\partial\mathcal{L}_{KD}}{\partial p_i^S} \frac{\partial p_i^S}{\partial z_i^S} + \sum_{\substack{k=1\\k\neq i}}^C \frac{\partial\mathcal{L}_{KD}}{\partial p_k^S} \frac{\partial p_k^S}{\partial z_i^S} \nonumber \\
    &= \left( - \frac{p_i^T}{p_i^S} \right) \left( p_i^S - (p_i^S)^2 \right) + \sum_{\substack{k=1\\k\neq i}}^C \left( - \frac{p_k^T}{p_k^S} \right) (-p_k^S \cdot p_i^S) \nonumber \\
    &= (p_i^T+\sum_{\substack{k=1\\k\neq i}}^Cp_k^T)p_i^S - p_i^T \nonumber \\[5pt]
    &= p_i^S - p_i^T,
\end{align}
as shown in Eq.~6 of the main paper.

\subsection{DKD (Eqs.~\ref{eq:dkd_partial_t} and \ref{eq:dkd_partial_nott})}
\label{sec:appendix_dkd}
The DKD loss is given by the weighted combination of the target class KD (TCKD) and the non-target class KD (NCKD):
\begin{align}
    \mathcal{L}_{DKD} = \alpha \mathcal{L}_{TCKD} + \beta \mathcal{L}_{NCKD},
\end{align}
as shown in Eq.~4 of the main paper.
The gradient of $\mathcal{L}_{DKD}$ with respect to $z_i^S$ is given by
\begin{align}
    \dfrac{\partial\mathcal{L}_{DKD}}{\partial z_i^S} &= \alpha \dfrac{\partial\mathcal{L}_{TCKD}}{\partial z_i^S} + \beta \dfrac{\partial\mathcal{L}_{NCKD}}{\partial z_i^S}.
    \label{eq:grad_dkd}
\end{align}


As in Eq.~7 of the main paper, $\mathcal{L}_{TCKD}$ is written as
\begin{align}
    \mathcal{L}_{TCKD} = p_t^T (\log p_t^T - \log p_t^S) + p_{\neg t}^T (\log p_{\neg t}^T - \log p_{\neg t}^S).
    \label{eq:loss_tckd}
\end{align}
Thus, 
\begin{align}
    \dfrac{\partial\mathcal{L}_{TCKD}}{\partial z_i^S} &= %
    \dfrac{\partial\mathcal{L}_{TCKD}}{\partial p_t^S}\dfrac{\partial p_t^S}{\partial z_i^S}+ \dfrac{\partial\mathcal{L}_{TCKD}}{\partial p_{\neg t}^S}\dfrac{\partial p_{\neg t}^S}{\partial z_i^S},
    \label{eq:grad_tckd}
\end{align}
where 
\begin{align}
    \dfrac{\partial\mathcal{L}_{TCKD}}{\partial p_t^S}=-\dfrac{p_t^T}{p_t^S}
    \label{eq:tckd_p1}
\end{align}
and
\begin{align}
    \dfrac{\partial\mathcal{L}_{TCKD}}{\partial p_{\neg t}^S}=-\dfrac{p_{\neg t}^T}{p_{\neg t}^S} 
    \label{eq:tckd_p2}
\end{align}
from Eq.~\ref{eq:loss_tckd}.

Let us consider two cases in computing Eq.~\ref{eq:grad_tckd}: $i=t$ and $i\neq t$.
First, when $i=t$, we need to obtain $\partial p_t^S / \partial z_t^S$ and $\partial p_{\neg t}^S / \partial z_t^S$.
They can be computed as
\begin{align}
    \dfrac{\partial p_t^S}{\partial z_t^S} &= %
    \dfrac{e^{z_t^S}\times\left(\sum_{l=1}^C e^{z_l^S}\right)-e^{z_t^S}\times e^{z_t^S}}{\left(\sum_{l=1}^C e^{z_l^S}\right)^2} \nonumber \\
    &= \dfrac{e^{z_t^S}}{\sum_{l=1}^C e^{z_l^S}} - \left( \dfrac{e^{z_t^S}}{\sum_{l=1}^C e^{z_l^S}} \right)^2 \nonumber \\[2pt]
    &=p_t^S-(p_t^S)^2 ,\label{eq:tckd_p_zt1}\\[5pt]
    \dfrac{\partial p_{\neg t}^S}{\partial z_t^S} &= %
    \dfrac{-\left(\sum_{m=1, m\neq t}^C e^{z_m^S}\right)\times e^{z_t^S}}{\left(\sum_{l=1}^C e^{z_l^S}\right)^2} \nonumber \\
    &= - \left( \dfrac{\sum_{m=1, m\neq t}^C e^{z_m^S}}{\sum_{l=1}^C e^{z_l^S}} \right) \left( \dfrac{e^{z_t^S}}{\sum_{l=1}^C e^{z_l^S}} \right) \nonumber \\[2pt]
    &=-p_{\neg t}^S\cdot p_t^S . \label{eq:tckd_p_zt2}
\end{align}
By inserting Eqs.~\ref{eq:tckd_p1} through \ref{eq:tckd_p_zt2} into Eq.~\ref{eq:grad_tckd}, we obtain 
\begin{align}
    \dfrac{\partial\mathcal{L}_{TCKD}}{\partial z_t^S} &= %
    \dfrac{\partial\mathcal{L}_{TCKD}}{\partial p_t^S}\dfrac{\partial p_t^S}{\partial z_t^S}+ \dfrac{\partial\mathcal{L}_{TCKD}}{\partial p_{\neg t}^S}\dfrac{\partial p_{\neg t}^S}{\partial z_t^S} \nonumber \\[3pt]
    &=\left(-\dfrac{p_t^T}{p_t^S}\right)\left(p_t^S-(p_t^S)^2\right) +\left(-\dfrac{p_{\neg t}^T}{p_{\neg t}^S}\right)\left(-p_{\neg t}^S\cdot p_t^S\right) \nonumber \\[3pt]
    &=(p_t^T+p_{\neg t}^T)p_t^S-p_t^T \nonumber \\[5pt]
    &=p_t^S-p_t^T.
    \label{eq:tckd_p_zt}
\end{align}

Second, for $i\neq t$, we compute $\partial p_t^S / \partial z_i^S$ and $\partial p_{\neg t}^S / \partial z_i^S$ as follows.
\begin{align}
    \dfrac{\partial p_t^S}{\partial z_i^S} &= %
    \dfrac{-e^{z_t^S}\times e^{z_i^S}}{\left(\sum_{l=1}^C e^{z_l^S}\right)^2} \nonumber \\
    &=-p_t^S\cdot p_i^S, \label{eq:tckd_p_znt1}\\[5pt]
    \dfrac{\partial p_{\neg t}^S}{\partial z_i^S} &= %
    \dfrac{e^{z_i^S}\times\left(\sum_{l=1}^C e^{z_l^S}\right)-\left(\sum_{m=1, m\neq t}^C e^{z_m^S}\right)\times e^{z_i^S}}{\left(\sum_{l=1}^C e^{z_l^S}\right)^2} \nonumber \\
    &=p_i^S-p_{\neg t}^Sp_i^S. \label{eq:tckd_p_znt2}
\end{align}
By inserting Eqs.~\ref{eq:tckd_p1}, \ref{eq:tckd_p2}, \ref{eq:tckd_p_znt1}, and \ref{eq:tckd_p_znt2} into Eq.~\ref{eq:grad_tckd}, we obtain 
\begin{align}
    \dfrac{\partial\mathcal{L}_{TCKD}}{\partial z_i^S} &= %
    \dfrac{\partial\mathcal{L}_{TCKD}}{\partial p_t^S}\dfrac{\partial p_t^S}{\partial z_i^S}+ \dfrac{\partial\mathcal{L}_{TCKD}}{\partial p_{\neg t}^S}\dfrac{\partial p_{\neg t}^S}{\partial z_i^S} \nonumber \\[3pt]
    &=\left(-\dfrac{p_t^T}{p_t^S}\right)\left(-p_t^S\cdot p_i^S\right) +\left(-\dfrac{p_{\neg t}^T}{p_{\neg t}^S}\right)\left(p_i^S-p_{\neg t}^Sp_i^S\right) \nonumber \\[3pt]
    &=\left\{\left(p_t^T+p_{\neg t}^T\right)-\dfrac{p_{\neg t}^T}{p_{\neg t}^S}\right\} p_i^S \nonumber \\[5pt]
    &=\left(1-\dfrac{p_{\neg t}^T}{p_{\neg t}^S}\right)p_i^S.
    \label{eq:grad_tckd_i}
\end{align}


$\mathcal{L}_{NCKD}$ is given by Eq.~\ref{eq:nckd} of the main paper, i.e.,
\begin{align}
    \mathcal{L}_{NCKD} = \sum_{\substack{i=1\\i\neq t}}^C \hat{p}_i^T (\log \hat{p}_i^T - \log \hat{p}_i^S).
    \label{eq:loss_nckd}
\end{align}
Since this is a function of $\hat{p}_i^S$, we have
\begin{align}
    \dfrac{\partial\mathcal{L}_{NCKD}}{\partial z_t^S} =0.
    \label{eq:grad_nckd_zt}
\end{align}
For $i\neq t$, we can write
\begin{align}
	\dfrac{\partial\mathcal{L}_{NCKD}}{\partial z_i^S} &=
    \sum_{k=1}^C \dfrac{\partial\mathcal{L}_{NCKD}}{\partial \hat{p}_k^S} \dfrac{\partial \hat{p}_k^S}{\partial z_i^S}.
    \label{eq:nckd_partial}
\end{align}
By differentiating Eq.~\ref{eq:loss_nckd}, we obtain
\begin{align}
    \dfrac{\partial\mathcal{L}_{NCKD}}{\partial \hat{p}_k^S}&=-\dfrac{\hat{p}_k^T}{\hat{p}_k^S}. \label{eq:nckd_p}
\end{align}
And, similarly to Eqs.~\ref{eq:kd_pi_zi} and \ref{eq:kd_pk_zi}, we can obtain
\begin{align}
    \dfrac{\partial\hat{p}_i^S}{\partial z_i^S} %
    &=\dfrac{ e^{z_i^S}\times\sum_{l=1, l\neq t}^C e^{z_l^S}-e^{z_i^S}\times e^{z_i^S}}{(\sum_{l=1, l\neq t}^C e^{z_l^S})^2} \nonumber \\[2pt]
    &=\hat{p}_i^S-(\hat{p}_i^S)^2, \label{eq:nckd_p_zt} \\[5pt]
    \dfrac{\partial\hat{p}_k^S}{\partial z_i^S} %
    &=\dfrac{ -e^{z_k^S}\times e^{z_i^S}}{(\sum_{l=1, l\neq t}^C e^{z_l^S})^2} \nonumber \\[2pt]
    &=-\hat{p}_k^S \cdot \hat{p}_i^S . \label{eq:nckd_p_znt}
\end{align}

By inserting Eqs.~\ref{eq:nckd_p}, \ref{eq:nckd_p_zt}, and \ref{eq:nckd_p_znt} into Eq.~\ref{eq:nckd_partial}, we get
{\footnotesize
\begin{align}
    \dfrac{\partial\mathcal{L}_{NCKD}}{\partial z_i^S} &=%
    \dfrac{\partial\mathcal{L}_{NCKD}}{\partial \hat{p}_i^S} \dfrac{\partial \hat{p}_i^S}{\partial z_i^S} + \sum_{\substack{k=1\\k\neq i}}^C \dfrac{\partial\mathcal{L}_{NCKD}}{\partial \hat{p}_k^S} \dfrac{\partial \hat{p}_k^S}{\partial z_i^S} \nonumber\\
    &= %
    \left(-\dfrac{\hat{p}_i^T}{\hat{p}_i^S}\right)\left(\hat{p}_i^S-(\hat{p}_i^S)^2\right) + \sum_{\substack{k=1\\k\neq i}}^C\left(-\dfrac{\hat{p}_k^T}{\hat{p}_k^S}\right)\left(-\hat{p}_k^S\hat{p}_i^S\right) \nonumber \\
    &= (\hat{p}_i^T+\sum_{\substack{k=1\\k\neq i}}^C\hat{p}_k^T)\hat{p}_i^S - \hat{p}_i^T \nonumber \\[3pt]
    &= \hat{p}_i^S-\hat{p}_i^T \nonumber \\[5pt]
    &= \dfrac{1}{p_{\neg t}^S}p_i^S-\dfrac{1}{p_{\neg t}^T}p_i^T. \label{eq:grad_nckd_znt}
\end{align}
}


Now, we are ready to compute Eq.~\ref{eq:grad_dkd}.
By inserting Eqs.~\ref{eq:tckd_p_zt}, \ref{eq:grad_tckd_i}, \ref{eq:grad_nckd_zt}, and \ref{eq:grad_nckd_znt} into Eq.~\ref{eq:grad_dkd}, we obtain
{\footnotesize
\begin{align}
\dfrac{\partial\mathcal{L}_{DKD}}{\partial z_i^S}%
    &=\begin{cases}
        \alpha(p_i^S - p_i^T) &(i=t) \\[5pt]
        \left\{ \alpha \left( 1- \dfrac{p_{\neg t}^T}{p_{\neg t}^S} \right) + \dfrac{\beta}{p_{\neg t}^S} \right\} p_i^S - \dfrac{\beta}{p_{\neg t}^T} p_i^T &(i\neq t)
    \end{cases}
    \nonumber
\end{align}
}

\subsection{AEKT (Eqs.~\ref{eq:grad_aekt_t} and \ref{eq:grad_aekt_nt})}
\label{sec:appendix_aekt}

The AEKT loss ($\mathcal{L}_{AEKT}$) is defined by
\begin{align}
    \mathcal{L}_{AEKT}&=\log{\left(\dfrac{p_t^T}{p_t^S}\right)}\cdot\left(1-2^{1-\frac{p_t^T}{\reallywideacute{p}_t^S}}\right),
    \label{eq:loss_aekt}
\end{align}
as in Eq.~\ref{eq:loss_AEKT} of the main paper.

Since $\mathcal{L}_{AEKT}$ is a function of $p_t^S$, its gradient with respect to $z_i^S$ can be written as
\begin{align}
	\frac{\partial \mathcal{L}_{AEKT}}{\partial z_i^S} &= \frac{\partial \mathcal{L}_{AEKT}}{\partial p_t^S} \frac{\partial p_t^S}{\partial z_i^S},
\end{align}
where
\begin{align}
    \dfrac{\partial\mathcal{L}_{AEKT}}{\partial p_t^S}&=-\dfrac{1}{p_t^S}\left(1-2^{1-\frac{p_t^T}{\reallywideacute{p}_t^S}}\right).
\end{align}
Using Eqs.~\ref{eq:tckd_p_zt1} and \ref{eq:tckd_p_znt1}, we can obtain
\begin{align}
    \dfrac{\partial\mathcal{L}_{AEKT}}{\partial z_i^S}
    &=\begin{cases}
        \left(- \dfrac{1}{p_t^S} \left(1-2^{1-\frac{p_t^T}{\reallywideacute{p}_t^S}} \right) \right) \left( p_t^S-(p_t^S)^2\right) & (i=t)  \\[15pt]
        \left(-\dfrac{1}{p_t^S} \left(1-2^{1-\frac{p_t^T}{\reallywideacute{p}_t^S}} \right) \right) \left( -p_t^S\cdot p_i^S \right) & (i \neq t)
    \end{cases} \nonumber \\[5pt]
    &=\begin{cases}
        -(1-p_t^S)\left(1-2^{1-\frac{p_t^T}{\reallywideacute{p}_t^S}}\right) & (i=t) \\[15pt]
        \left(1-2^{1-\frac{p_t^T}{\reallywideacute{p}_t^S}}\right)p_i^S & (i \neq t)
    \end{cases}
\end{align}

\section{$1-p_t^S$ in Eq.~\ref{eq:total_partial_t}}
\begin{figure}[hbt!]
    \centering
    \begin{subfigure}{0.45\textwidth}
        \includegraphics[width=\textwidth]{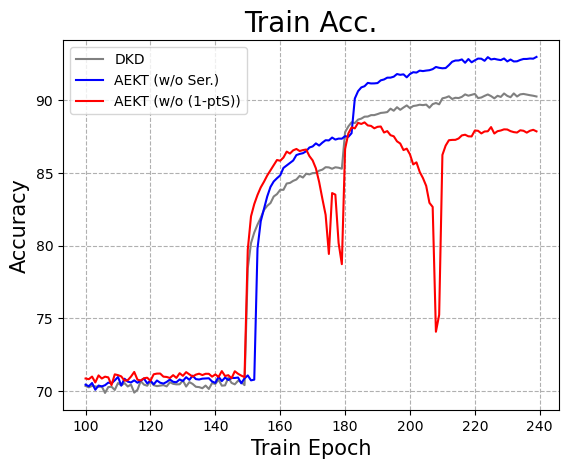}
    \end{subfigure}
    \hfill
    \begin{subfigure}{0.45\textwidth}
        \includegraphics[width=\textwidth]{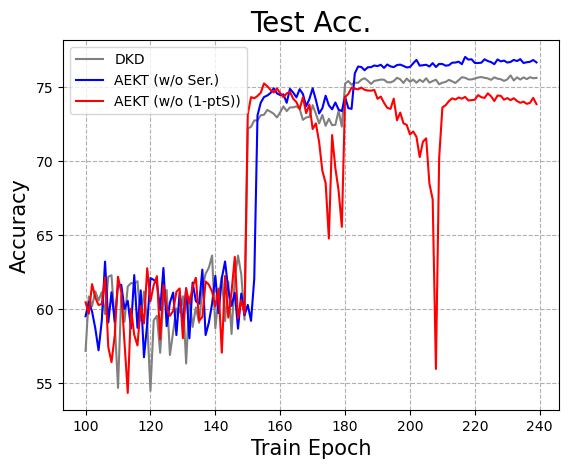}
    \end{subfigure}
    \caption{\textbf{Accuracy comparison with and without the $1-p_t^S$ term in the gradient backpropagated to the student model's target logit}}
    \label{fig:grad_diff}
\end{figure}

In Eq.~\ref{eq:total_partial_t} of the main paper, the newly introduced term contains $1-p_t^S$, of which effects were explained in the main paper.
Here, we also verify the importance of this term experimentally.
In other words, we deliberately replace this term with 1 during training and observe that the performance of the proposed method is degraded, even worse than that of DKD, as shown in Fig.~\ref{fig:grad_diff}.

\section{Experimental settings}
All experiments are conducted using PyTorch \cite{pytorch} with NVIDIA RTX 4090 and A6000 GPUs. The reported results are averages over three trials.
\label{sec:appendix_experimental_setting}
\subsection{CIFAR-100}
The student models are trained for 240 epochs. The SGD optimizer is used with a weight decay parameter of $5\times10^{-4}$ and a momentum parameter of 0.9. The batch size is set to 64. The learning rate is reduced by 1/10 every 30 epochs from 150 epochs. The initial learning rate is set to 0.05 for distillation between the same architecture and 0.01 for distillation between different architectures. The temperature for the softmax operation is set to 4. And we also utilize 20-epoch linear warm-up for fair comparison with DKD.

\begin{table}[t]
    \begin{subtable}[h]{\textwidth}
        \centering
        \begin{tabular}{c|c|c|c}
            \Xhline{2pt}
            Teacher   & $\alpha$             & $\beta$              & \textbf{$\gamma$} \\
            \Xhline{2pt}
            ResNet56  & \multirow{6}{*}{1.0} & \multirow{2}{*}{2.0} & \multirow{2}{*}{0.25} \\
            ResNet110 &                      &                      & \\
            \cline{3-4}
            WRN-40-2  &                      & \multirow{2}{*}{6.0} & \multirow{4}{*}{0.5} \\
            VGG13     &                      &                      & \\
            \cline{3-3}
            ResNet50  &                      & \multirow{2}{*}{8.0} & \\
            ResNet32x4&                      &                      & \\
            \Xhline{2pt}
        \end{tabular}
        \caption{\textbf{AEKT w/o Ser. for all cases.}}
        \label{tab:appendix_hyperparameter_CIFAR100_wo_ser}
    \end{subtable}
    \\
    
    \begin{subtable}[h]{0.48\textwidth}
        \centering
        \begin{tabular}{c|c|c|c}
            \Xhline{2pt}
            Teacher   & $\alpha$             & $\beta$              & \textbf{$\gamma$} \\
            \Xhline{2pt}
            ResNet56  & \multirow{6}{*}{0.5} & \multirow{2}{*}{1.0} & \multirow{2}{*}{0.05} \\
            ResNet110 &                      &                      & \\
            \cline{3-4}
            WRN-40-2  &                      & \multirow{2}{*}{3.0} & \multirow{4}{*}{0.1} \\
            VGG13     &                      &                      & \\
            \cline{3-3}
            ResNet32x4&                      & 4.0                  & \\
            \Xhline{2pt}
        \end{tabular}
        \caption{\textbf{AEKT for same architecture cases.}}
        \label{tab:appendix_hyperparameter_CIFAR100_same}
    \end{subtable}
    \begin{subtable}[h]{0.48\textwidth}
        \centering
        \begin{tabular}{c|c|c|c}
            \Xhline{2pt}
            Teacher   & $\alpha$             & $\beta$              & \textbf{$\gamma$} \\
            \Xhline{2pt}
            WRN-40-2  & \multirow{4}{*}{0.5} & \multirow{2}{*}{6.0} & \multirow{4}{*}{0.1} \\
            VGG13     &                      &                      & \\
            \cline{3-3}
            ResNet50  &                      & \multirow{2}{*}{8.0} & \\
            ResNet32x4&                      &                      & \\
            \Xhline{2pt}
        \end{tabular}
        \caption{\textbf{AEKT for different architecture cases.}}
        \label{tab:appendix_hyperparameter_CIFAR100_diff}
    \end{subtable}
    \caption{\textbf{Hyperparameter settings for CIFAR-100}}
\end{table}


The weight for the cross-entropy loss is always set to 1.0. In DKD, the values of $\alpha$ and $\beta$ are set to vary depending on the teacher-student model pair according to a guidance rule, which we also follow, as shown in \autoref{tab:appendix_hyperparameter_CIFAR100_wo_ser}. When the task serialization is not used, we set $\gamma$ as 0.25 for ResNet56$\rightarrow$ResNet20 and ResNet110$\rightarrow$ResNet32 and 0.5 for all other cases. With task serialization, we set the hyperparameters as shown in \autoref{tab:appendix_hyperparameter_CIFAR100_same} for same architecture cases, and \autoref{tab:appendix_hyperparameter_CIFAR100_diff} for different architecture cases.

\subsubsection{Hyperparameter $\gamma$ without task serialization}
In DKD, the value of $\beta$ is set to be larger when a more powerful teacher model showing high confidence is used, which promotes adaptive learning of non-target class information. In the same vein, we set the value of $\gamma$ to be correlated with that of $\beta$ so that adaptive learning of target class information is also promoted for a powerful teacher model.

\subsubsection{Hyperparameter with task serialization}\label{sec:appendix_experimental_setting_serial}
When task serialization is used, the classification logits and distillation logits are decoupled, and the learning of the student model may focus more on either the classification task or the distillation task.
In order to sufficiently focus on the classification task and ensure high classification performance, we decrease all three hyperparameters as shown in \autoref{tab:appendix_hyperparameter_CIFAR100_same} so that the learning can focus more on classification.
In different architecture cases, on the other hand, we found that using a relatively large value of $\beta$ is beneficial, as shown in \autoref{tab:appendix_additional_experiment}. It is thought that promoting the information about the non-target classes is important for effective distillation between different architecture pairs.



\begin{table}[h]
        \centering
        \begin{tabular}{c|c|c|c|cc|c||c}
            \Xhline{2pt}
            \multicolumn{2}{c|}{Model} & \multicolumn{2}{c|}{Baseline} & \multicolumn{3}{c||}{Hyperparam.} & \multirow{2}{*}{Acc.} \\
            \cline{1-7}
            Teacher   & Student       & Teacher & Student            & $\alpha$ & $\gamma$ & $\beta$ & \\
            \Xhline{2pt}
            
            \multirow{2}{*}{ResNet32x4} & \multirow{2}{*}{WRN-40-2} & \multirow{2}{*}{79.42} & \multirow{2}{*}{75.61} & \multirow{2}{*}{0.5} & \multirow{2}{*}{0.1} & \multirow{1}{*}{8.0} & \textbf{78.12} \\
            \cline{7-8}
            &&&&&& \multirow{1}{*}{4.0} & 77.92 \\
            
            \hline
            
            \multirow{2}{*}{ResNet32x4} & \multirow{2}{*}{ResNet50} & \multirow{2}{*}{79.42} & \multirow{2}{*}{79.34} & \multirow{2}{*}{0.5} & \multirow{2}{*}{0.1}  & \multirow{1}{*}{8.0} & \textbf{81.18} \\
            \cline{7-8}
            &&&&&& \multirow{1}{*}{4.0} & 80.99 \\
            
            \hline

            \multirow{2}{*}{WRN-40-2} & \multirow{2}{*}{VGG13} & \multirow{2}{*}{75.61} & \multirow{2}{*}{74.64} & \multirow{2}{*}{0.5} & \multirow{2}{*}{0.1}  & \multirow{1}{*}{6.0} & \textbf{77.83} \\
            \cline{7-8}
            &&&&&& \multirow{1}{*}{3.0} & 77.62 \\
            
            \Xhline{2pt}
        \end{tabular}
        \caption{\textbf{Additional experiments for different architecture pairs of teachers and students.}}
        \label{tab:appendix_additional_experiment}
\end{table}
\begin{table}[h]
    \centering
        \begin{tabular}{c|c|c|c}
            \Xhline{2pt}
            Teacher   & $\alpha$             & $\beta$              & $\gamma$\\
            \Xhline{2pt}
            DeiT-Tiny & \multirow{4}{*}{1.0} & \multirow{2}{*}{1.0} & \multirow{4}{*}{0.25}\\
            ResNet34  &                      &                      & \\
            \cline{3-3}
            ResNet50  &                      & \multirow{2}{*}{2.0} & \\
            Swin-Tiny &                      &                      & \\
            \Xhline{2pt}
        \end{tabular}

    \caption{\textbf{Hyperparameter settings for ImageNet.}}
    \label{tab:appendix_hyperparameter_IMAGENET}
\end{table}

\subsection{ImageNet}
The student models are trained using the SGD optimizer with a weight decay parameter of $10^{-4}$ and a momentum parameter of 0.9 for 100 epochs. The learning rate is reduced by 1/10 at every 30 epochs. The initial learning rate is set to 0.1. The batch size is set to 512. The values of ($\alpha$, $\beta$, $\gamma$) are set as shown in \autoref{tab:appendix_hyperparameter_IMAGENET}. The temperature is set to 1. The weight for the cross-entropy loss is set to 1.0.

In the case of ImageNet, we found that the same hyperparameter values can be used for both with and without task serialization. This is probably because due the large number of classes, the learning is hardly biased to the distillation task and thus the hyperparameter control used in Section \ref{sec:appendix_experimental_setting_serial} is not necessary.

\section{Pseudo code of AEKT in a PyTorch style}

\label{sec:appendix_pseudo_code}
\begin{lstlisting}[language=Python, basicstyle=\small]
# l_T_c : Output logits of Teacher model
# l_S_c : Output logits of Student model for classification
# l_S_d : Output logits of Student model for distillation
# Linear : FC layer for task serialization
# t : target (N*C, bool)
# T : Temperature for distillation
# alpha, beta, gamma : hyperparameter for AEKT

l_T_c = teacher(image)
l_S_c = student(image)

l_S_d = Linear(l_S_c)   # Task Serialization

p_T = softmax(l_T_c/T)
p_S = softmax(l_S_d/T)

# Eq. A.2 and A.3
p_t_T, p_nt_sum_T = p_T[t], p_T[1-t].sum(1)
p_t_S, p_nt_sum_S = p_S[t], p_S[1-t].sum(1)

# Eq. A.4
p_nt_T = softmax(l_T_c[1-t]/T)
p_nt_S = softmax(l_S_d[1-t]/T)

p_t_S_2 = p_t_s.clone().detach()

# Loss TCKD
TCKD = kl_div(log(p_t_S), p_t_T) + kl_div(log(p_nt_sum_S), p_nt_sum_T)
# Loss NCKD
NCKD = kl_div(log(p_nt_S), p_nt_T)
# Loss AEKT
AEKT = (log(p_t_T) - log(p_t_S)) * (1 - 2**(1 - p_t_T/p_t_S_2))

# Total loss of our proposed method AEKT, Eq. 16
TOTAL = (alpha * TCKD + beta * NCKD + gamma * AEKT) * T**2
\end{lstlisting}

\section{Discussion}
\label{sec:appendix_discussion}
\paragraph{Limitations.}
Our method can perform even better via further parameter tuning as shown in Section \ref{sec:Experiments_ImageNet}. A more effective parameter tuning method will be a topic of our future work. 
\paragraph{Societal impacts.}
As mentioned above, finding better-performing hyperparameters may be resource-intensive.
Therefore, research into methods for selecting optimal hyperparameters will be beneficial.
If this issue is addressed, our method can enable the use of compact, high-performing models in resource-constrained environments, such as edge devices, and reduce energy consumption.

\end{document}